\documentclass[fleqn,10pt,twocolumn]{ICCAS2020}
\usepackage{makecell}
\usepackage{multirow}
\usepackage[table]{xcolor}
\usepackage{caption}
\usepackage{adjustbox}	
\usepackage{algorithm}
\usepackage[noend]{algpseudocode}

\definecolor{LightCyan}{rgb}{0.88,1,1}
\definecolor{White}{rgb}{1,1,1} 
\definecolor{CuGray}{gray}{0.97}
\newcolumntype{g}{>{\columncolor{CuGray}}c}

\def\eg{\emph{e.g}.,~}

\newcommand{\figref}[1]{Fig.~\ref{#1}}
\newcommand{\tabref}[1]{Table~\ref{#1}}

\begin{document}

\title{Retaining Image Feature Matching Performance Under Low Light Conditions}

\author{Pranjay Shyam${}^{*}$, Antyanta Bangunharcana${}^{*}$ and Kyung-Soo Kim ${}$}

\affils{ ${}$Department of Mechanical Engineering\\ Korea Advanced Institute of Science and Technology (KAIST), \\
Daejeon, 34141, Republic of Korea.\\ \{pranjayshyam, antabangun, kyungsookim\}@kaist.ac.kr }

\thanks{ \noindent
  * Denotes Equal Contribution \\
  This paper was supported by Korea Civil-Military Combined Technology Development Project (Task No. 19-CMGU-02).
 }

\abstract{
Poor image quality in low light images may result in a reduced number of feature matching between images. In this paper, we investigate the performance of feature extraction algorithms in low light environments. To find an optimal setting to retain feature matching performance in low light images, we look into the effect of changing feature acceptance threshold for feature detector and adding pre-processing in the form of Low Light Image Enhancement (LLIE) prior to feature detection. We observe that even in low light images, feature matching using traditional hand-crafted feature detectors still performs reasonably well by lowering the threshold parameter. We also show that applying Low Light Image Enhancement (LLIE) algorithms can improve feature matching even more when paired with the right feature extraction algorithm.
}


\maketitle

%-----------------------------------------------------------------------

\section{Introduction}

% BEGIN WITH FEATURE MATCHING
Feature matching between image pair is a building block for high level tasks such as Simultaneous Localization and Mapping (SLAM) \cite{mur2015orb,mur2017orb}, Image alignment\cite{szeliski2006image},  3D Reconstruction. It relies on image feature extractors that detects feature points and compute the corresponding descriptors. Some of the well known hand-crafted feature extractors are SIFT \cite{lowe2004distinctive}, SURF \cite{bay2006surf}, ORB \cite{rublee2011orb}, AKAZE \cite{alcantarilla2011fast}, and BRISK \cite{leutenegger2011brisk}. These hand-crafted features often perform robustly in properly illuminated conditions. However, in real world scenarios illumination variations affect image quality by changing color distribution of objects captured within an image, adversely affecting the performance of underlying image processing algorithms that rely on feature matching techniques. 

Of late deep learning based algorithms have demonstrated state of the art performance in various tasks, with deep learning based image feature extractor \cite{detone2018superpoint} and matcher \cite{sarlin2020superglue} gaining popularity. To retain algorithm performance under diverse illumination conditions different works \cite{michaelis2019benchmarking} recommended extending the training dataset for deep learning based algorithms to account for scenarios exhibiting extreme illumination variations using real \cite{loh2019getting} and synthetic \cite{dai2018dark,liu2020lane} samples. However, this requires collecting, aligning and training on more low light image pairs, which is a time consuming and expensive process. Thus even until now, hand crafted features are still relied upon to perform feature matching.

% WHAT WE ARE TRYING TO WORK OUT
On the other hand, performance of traditional hand-crafted feature extractors cannot be guaranteed in low light conditions, primarily since they were designed to work with illuminated images. This hinders utilization of such techniques in applications wherein deployment conditions pertain to low light scenarios \eg{moon rovers, disaster response robot}. A simple approach to address this issue is to enhance the low light image as a preprocessing step, prior to feature extraction via enhancement mechanism that have been extensively studied. However directly using CNN based enhancement techniques doesn't guarantee performance improvement, primarily due to noise amplification and image stylization that increase number of extracted features but these features couldn't be matched within the image pairs. Furthermore, lowering the threshold of feature matching algorithms generally improve the matching quality. We summarize our contributions as,
\begin{itemize}
    \item We study the matching performance of handcrafted feature extractors in low light images with varying thresholds.
    \item We analyze the effects of integrating image enhancement algorithms on performance of feature matching on our test dataset. 
    \item We also compare the enhancement results to determine the best approach for enhancing images while minimizing the noisy pixels.  
\end{itemize}

To the best of our knowledge, no study on feature extraction performance in low light condition has been performed before. We hope that this study would improve future works on feature matching based computer vision tasks in low light environment.

\section{Related Works}
\subsection{Feature Matching}
 Harris corner detector \cite{harris1988combined} was one of the earliest work on feature extraction. Shi-Tomasi proposed a modified the scoring to extract corners in GFTT \cite{shi1994good}. Lowe proposed SIFT \cite{lowe2004distinctive} to detect features of multiple scales with Gaussian scale space on an image and extract rotation invariant descriptors. It remains one of the most popular choice for feature matching due to its robustness, but with a high computational cost. SURF \cite{bay2006surf} improved on the speed of feature detection using integral images. These two however, extracts descriptors with floating point numbers, resulting in a high computation cost for feature matching.

\begin{figure*}[!ht]
    \centering
    \begin{tabular}{c c}
    \includegraphics[width=\columnwidth]{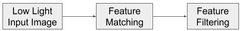} & 
    \includegraphics[width=\columnwidth, height=1.0cm]{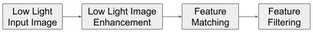} \\
    (a) & (b) \\
    \end{tabular}
\caption{(a) Standard and (b) Modified feature extraction pipeline by Integrating deep learning based light enhancement algorithms.}
\label{fig:setup}
\end{figure*}

Various binary descriptors are proposed to improve the feature matching speed for real-time application. BRIEF\cite{calonder2010brief} presents an efficient binary descriptor. ORB \cite{rublee2011orb} expand on this by modifying it to be rotation invariant as well as modifying a fast feature detector called FAST \cite{rosten2006machine}. It became popular due to its successful application in real-time SLAM \cite{mur2015orb,mur2017orb}. 
BRISK \cite{leutenegger2011brisk} is another efficient algorithm that utilize FAST-based detector and binary descriptor. 

Unlike SIFT, KAZE \cite{alcantarilla2012kaze} builds scale space in a non-linear manner to better retain object boundaries in images instead of smoothing them with Gaussian filter. AKAZE \cite{alcantarilla2011fast} is a faster and more efficient improvement on KAZE. Previous research \cite{tareen2018comparative} compared the performance of some of the mentioned algorithms.

\subsection{Low Light Image Enhancement}
Improving illumination quality of an image is a long researched topic with classical approaches such as histogram equalization \cite{pisano1998contrast,lee2013contrast} and gamma correction \cite{huang2012efficient} focusing on improving contrast of the complete image, ignoring region specific enhancement which leads to over and under saturation of regions within an image. To improve the performance of such systems, Retinex theory \cite{land1977retinex} was proposed wherein an image is represented into reflectance and illumination maps that represent the color and lighting information respectively. Leveraging the feature extraction capabilities of CNNs, different works were able to enhance image quality at both local and global level using a large labelled set of paired images. Specifically, MBLLEN\cite{Lv2018MBLLEN} proposed a multi-branch enhancement network to extract and enrich features across multiple networks for improving illumination condition within an image. GLADNet \cite{wang2018gladnet}, functions by first estimating a global illumination map and subsequently performs detail reconstruction to recover features during downsampling. KinD\cite{zhang2019kindling} and KinD++\cite{zhang2019kindling} follow retinex theory to decompose impose into illumination and reflectance maps and estimate these maps concurrently following a supervised learning approach, whereas RetinexNet\cite{Chen2018Retinex} followed similar principle and first decomposes the image into reflectance and illumination maps, enhances them and subsequently reconstructs enhanced image using improved maps.

\section{Methodologies}
% BRIEF INTRO
In this section, we describe the procedure of our experiments to study the performance of various feature extractors in raw and enhanced low light images. 

\subsection{Feature Matching}

We analyze the performance of feature extractor of images captured in low illumination settings, by extracting feature points and matching the features between image pairs (Fig. \ref{fig:setup}(a)). However, such an approach doesn't ensure correct feature matching thus, we utilize homography between image pairs to obtain inliers and use them to filter the extracted features. 

For this study, the feature detector-descriptors we are investigating are SIFT \cite{lowe2004distinctive}, SURF \cite{bay2006surf}, ORB \cite{rublee2011orb}, AKAZE \cite{alcantarilla2011fast}, BRISK \cite{leutenegger2011brisk}, and GFTT \cite{shi1994good} detector paired with BRIEF \cite{calonder2010brief} descriptor. To match the features, we follow the Nearest Neighbor Distance Ratio (NNDR) method by accepting good matches when the distance of the closest matching candidate is more than 0.7 times the distance of the second matching candidate. L2-norm is used to compute matching distance for SIFT and SURF, and Hamming distance is used for the other binary-type descriptors.

From the obtained feature matches, we find the matching inliers using RANSAC to compute homography transformation of the image pair. A match is rejected if the re-projection error of the homography transformed points is larger than $10.0$. 

Each of the aforementioned feature extractors have some form of threshold parameter to accept or reject candidate points for feature detection based on the feature strength. In low light images, it is to be expected that the strength of features are lower compared to illuminated images. By lowering feature acceptance thresholds we may improve detection rate. However, the hand-crafted feature descriptors are aimed towards illuminated image features, so the subsequent feature matching may not work in spite of the increased number of extracted features. For that reason, we investigate the performance of the feature extractor algorithms with lowered thresholds in this paper.

\subsection{Dataset Description}

\begin{figure*}[ht]
    \centering
    \begin{tabular}{c c}
    \includegraphics[width=0.23\linewidth]{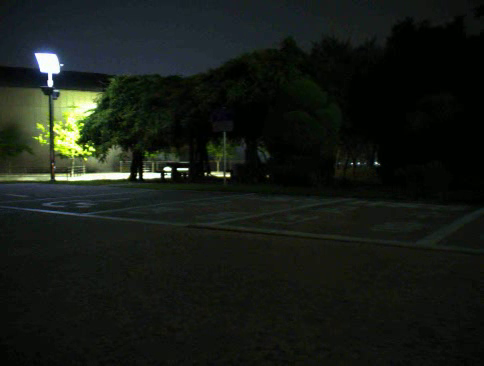}
    \includegraphics[width=0.23\linewidth]{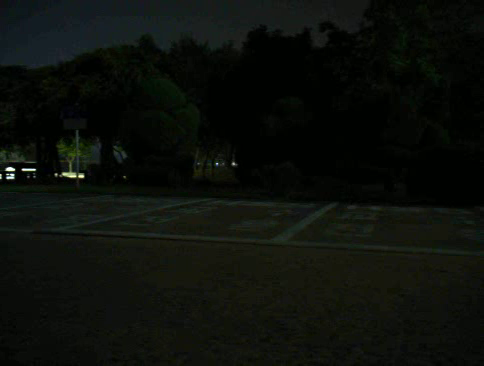} &
    \includegraphics[width=0.23\linewidth]{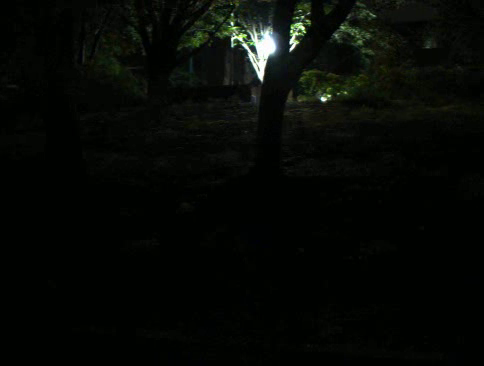}  
    \includegraphics[width=0.23\linewidth]{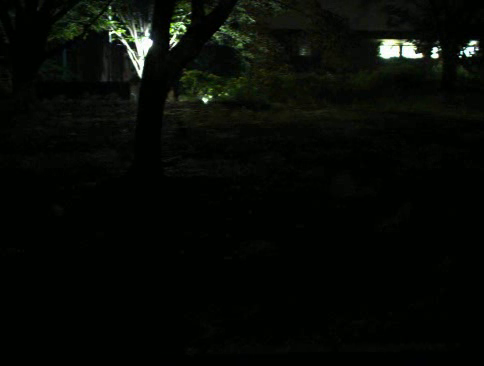} \\
    \multicolumn{2}{c}{Outdoor Locations} \\
    \includegraphics[width=0.23\linewidth]{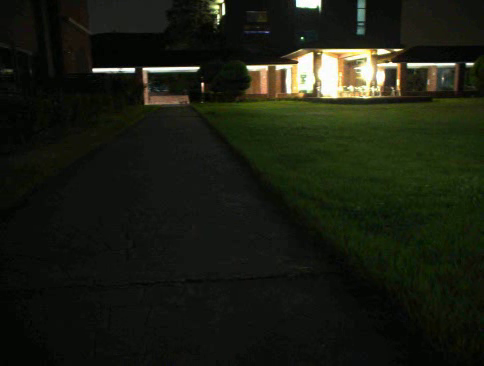}  
    \includegraphics[width=0.23\linewidth]{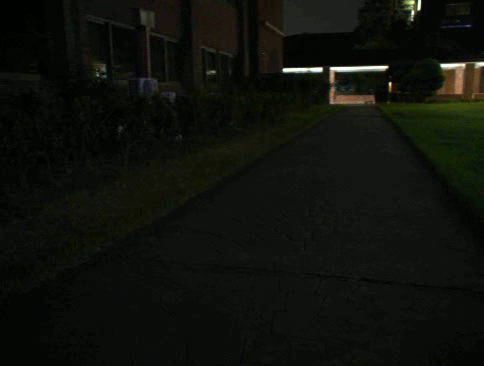} &
    \includegraphics[width=0.23\linewidth]{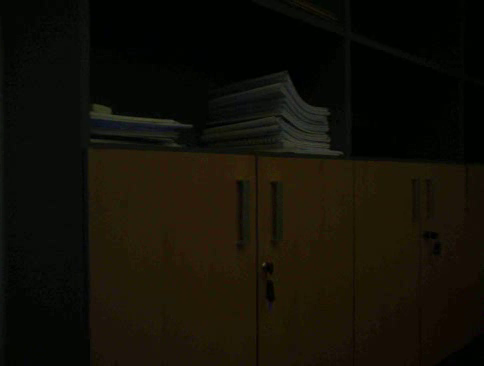}
    \includegraphics[width=0.23\linewidth]{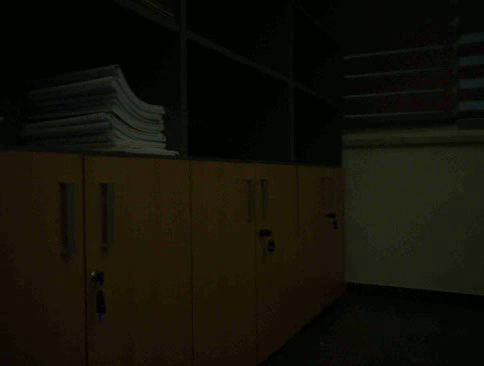} \\  
    Outdoor Location & Indoor Location \\
    \end{tabular}
% \captionsetup{justification=centering,margin=1cm}
\caption{Low Light Image pairs captured by rotating camera at different locations, highlighting local illumination sources and its effect on color distribution of objects within an image.}
\label{fig:dataset}
\end{figure*}
We collected 4 sets of image pairs captured in indoor (office room) and outdoor (parking lot, field, alley) low light environments with each set containing more than 5 image pairs. To obtain image pairs related by homography, the second image of each pair is captured by only rotating the camera after the first image is captured. \figref{fig:dataset} shows a examples of captured image pairs in each of the 4 sets. 

\subsection{Low Light Image Enhancement}
To determine if pairing image enhancement networks would improve the performance of feature matching, we select different CNN based enhancement algorithms such as MBLLEN\cite{Lv2018MBLLEN}, GLADNet \cite{wang2018gladnet}, KinD\cite{zhang2019kindling}, KinD++\cite{zhang2019kindling} and RetinexNet\cite{Chen2018Retinex}. Our motivation of using CNN based image enhancement techniques originates from its superior performance on different datasets, arising from its ability to non linearly enhance local regions. In this study, we retrain these algorithms on LOL dataset \cite{Chen2018Retinex} that comprises of 500 image pairs, divided into 485 training and 15 test image pairs and evaluate its performance using peak-signal-to-noise-ratio (PSNR) and Structural Similarity (SSIM)\cite{wang2004image} metrics, the results of which are summarized in \tabref{tab:tab_2}.

\begin{table}[h]
    \centering
    \setlength{\tabcolsep}{12pt}
    \caption{Evaluation of different low light image enhancement algorithms on LOL dataset}
    \begin{tabular}{c|c|c}
	\Xhline{3\arrayrulewidth}
    Algorithm & PSNR & SSIM \\
	\Xhline{2\arrayrulewidth}
    \hline
    \rowcolor{CuGray}
    Linear & 12.1706 & 0.5604\\
    MBLLEN\cite{Lv2018MBLLEN} & 17.8583 & 0.7247\\
    \rowcolor{CuGray}
    GLADNet \cite{wang2018gladnet} & 19.7182 & 0.6820\\
    KinD\cite{zhang2019kindling} & 17.6476 & 0.7715 \\
    \rowcolor{CuGray}
    KinD++\cite{zhang2019kindling} & 17.7518 & 0.7581 \\
    RetinexNet\cite{Chen2018Retinex} & 16.7740 & 0.4250\\
	\Xhline{3\arrayrulewidth}
    \end{tabular}
    \label{tab:tab_2}
\end{table}

\section{Experiments}
In this section, we first evaluate the performance of different CNN based enhancement techniques, focusing on noise amplification and image stylization and subsequently evaluate the performance of feature extractors by comparing the average number across our dataset of features extracted, matched features along with the number of accepted inliers in low light image pairs.

The extensive results of our feature matching experiments is shown in \tabref{table:featmatch}. For each feature extractors, we show the matching results performed directly in low light images as well as in LLIE processed images. We also show the results with multiple different feature detector threshold parameters. It is important to note that the left most threshold value in each feature extractor is the default value in the OpenCV implementation.

\begin{figure}[ht]
    \centering
    \begin{tabular}{c c}
    \includegraphics[width=0.475\linewidth]{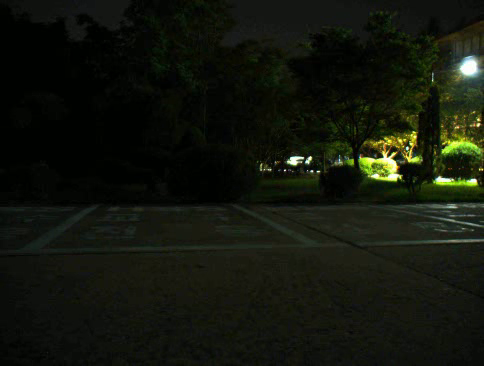} &
    \includegraphics[width=0.475\linewidth]{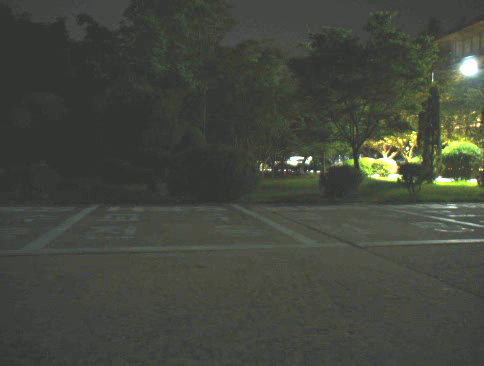} \\
    (a) & (b) \\
    \includegraphics[width=0.475\linewidth]{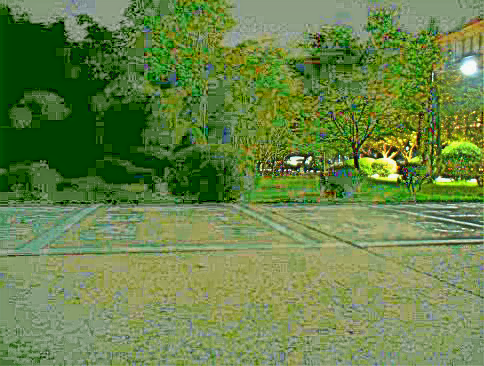} &
    \includegraphics[width=0.475\linewidth]{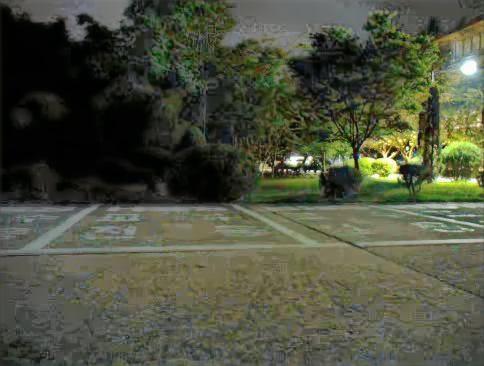} \\
    (c) & (d) \\
    \includegraphics[width=0.475\linewidth]{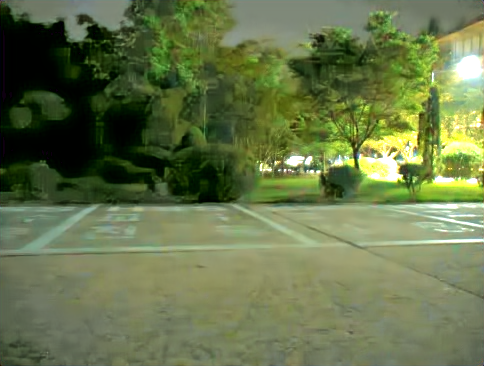} &
    \includegraphics[width=0.475\linewidth]{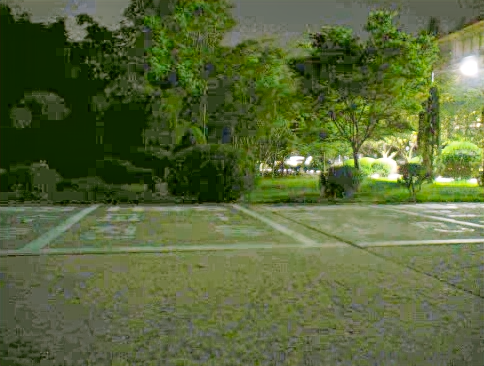} \\
    (e) & (f) \\
    \end{tabular}
% \captionsetup{justification=centering,margin=1cm}
\caption{Qualitative results from different enhancement networks for a given (a) low light image and corresponding results from (b) Linear Enhancement, (c) RetinexNet, (d) Kind, (e) MBLLEN and (f) GLAD.}
\label{fig:llie_sample}
\end{figure}

We observe that at higher threshold values, excluding linear based image enhancement, CNN based enhancement techniques improves the number of detected as well as the matched features. This is to be expected as image enhancement strengthen the edges and corners within the image as shown in \figref{fig:llie_sample}. However, simply lowering the feature acceptance threshold consistently improve performance massively in raw images. This improvement is not seen as much in processed images as most of the features were already strengthened in the first place. Moreover, adding image enhancement prior to feature extraction with lower thresholds does not uniformly improve performance for all feature extractors.

% \subsection{Analysis of different low light enhancement techniques}

From \tabref{table:featmatch}, we infer that at lower thresholds, integrating enhancement networks such as KinD and MBLLEN improves the feature matching performance of AKAZE and BRISK, however it reduces the performance of SURF and ORB while demonstrating negligible difference in performance of SIFT and GFTT-BRIEF. We attribute this inconsistency in performance arising from inconsistent introduction of new features after application of enhancement techniques, as demonstrated in \figref{fig:llie_sample}. While linear enhancement focuses on improving global contrast, different CNN based techniques introduce different categories of artifacts. Specifically, RetinexNet (\figref{fig:llie_sample} (c)) improves illumination performance but it also generates stylized features thereby destroying the natural features present in the image. Similarly GLAD affects the natural image features by introducing large pixelated noisy features arising mainly due to its inability to extracting and represent features from small pixels. On contrary, MBLLEN and KinD performs image enhancement without introducing a significant amount of noisy pixels or stylizing images which helps in improving the performance of underlying feature extraction and matching stage. Due to this side effect of image enhancement and the different ways each feature extractors detect and obtain descriptors, some image enhancement and feature extractor pairs works well while some others doesn't.

% \subsection{Analysis of integrating image enhancement with feature matching}

From this observation, in resource constrained application such as SLAM in mobile robotics or Augmented Reality in mobile phones where the use of computationally demanding deep learning techniques is impossible, we recommend a mere adjustment in threshold parameters to improve feature matching. However, image enhancement technique can be applied if paired with the correct feature extractors if a further improvement upon the feature matching is desired. Out of all the feature extractors we experimented on, AKAZE seems to perform the best in low light environment followed by BRISK, with MBLLEN and KinD pairing well with them.

% Some analysis points to make:
% >> Lowering threshold improves performance massively in raw images but not as much in processed images

% >> A lot more features are extracted from RetinexNet processed image but not matched well, a sign of the algorithm destroying natural image features

\begin{table*}
\begin{center}
\caption{Average number of extracted and matched features as well as accepted inliers based on homography fitting for various feature detector-descriptor with multiple thresholds on low light images enhanced with several LLIE algorithms. \underline{Underlined} numbers represents the highest numbers for each feature extractor. \textbf{Bold} numbers are the overall highest numbers.}
\label{table:featmatch}
\begin{tabular}{c|g|c|c|c}
	\Xhline{3\arrayrulewidth}
	Feature detector-descriptor & LLIE algorithm & \# Features detected & \# Features matched & \# Inliers accepted \\  
	\Xhline{2\arrayrulewidth}
	\hline
% 	\Xhline{2\arrayrulewidth}
	& & \multicolumn{3}{c}{Contrast Threshold : 0.04 / 0.01 / 0.001} \\
	\cline{3-5}
	\multirow{7}{*}{SIFT} & Raw & 61 / 584 / 2461 & 25 / 73 / 84 & 21 / 67 / 73 \\
	& Linear & 62 / 585 / 2463 & 24 / 72 / 83 & 21 / 66 / 72 \\
	& MBLLEN & 873 / 1863 / 2856 & 68 / 82 / 91 & 63 / 70 / 69 \\
	& GLAD & 1137 / 2475 / \underline{2932} & 60 / 68 / 73 & 54 / 57 / 57 \\
	& KinD & 672 / 2304 / 2881 & 68 / 78 / 82 & 63 / 70 / 70 \\
	& KinDpp & 1619 / 2691 / 3152 & 77 / 81 / \underline{101} & 72 / 72 / \underline{74} \\
	& RetinexNet & 1635 / 2588 / 2878 & 38 / 39 / 43 & 34 / 34 / 34 \\
	\Xhline{2\arrayrulewidth}
	\hline
	& & \multicolumn{3}{c}{Hessian Threshold : 100 / 10 / 1} \\
	\cline{3-5}
	\multirow{7}{*}{SURF} & Raw & 123 / 650 / 1514 & 42 / 87 / \underline{102} & 37 / 76 / 85 \\
	& Linear & 123 / 651 / 1513 & 41 / 87 / 101 & 36 / 75 / \underline{86} \\
	& MBLLEN & 859 / 1249 / 1427 & 82 / 92 / 95 & 71 / 75 / 76 \\
	& GLAD & 1065 / 1721 / 1967 & 66 / 70 / 72 & 55 / 57 / 57 \\
	& KinD & 837 / 1393 / 1612 & 73 / 81 / 83 & 64 / 69 / 69 \\
	& KinDpp & 1271 / 1702 / 1871 & 71 / 74 / 75 & 60 / 62 / 62 \\
	& RetinexNet & 1465 / 2404 / \underline{2744} & 40 / 45 / 48 & 31 / 33 / 33 \\
	\Xhline{2\arrayrulewidth}
	\hline
	& & \multicolumn{3}{c}{FAST Threshold : 20 / 2} \\
	\cline{3-5}
	\multirow{7}{*}{ORB} & Raw & 175 / 481 & 38 / 65 & 35 / 61 \\
	& Linear & 173 / 481 & 38 / \underline{66} & 36 / \underline{62} \\
	& MBLLEN & 456 / \underline{487} & 54 / 55 & 51 / 52 \\
	& GLAD & 464 / \underline{487} & 47 / 47 & 43 / 44 \\
	& KinD & 450 / \underline{487} & 61 / 62 & 59 / 59 \\
	& KinDpp & 476 / \underline{487} & 54 / 54 & 50 / 50 \\
	& RetinexNet & 469 / \underline{487} & 27 / 28 & 25 / 24 \\
	\Xhline{2\arrayrulewidth}
	\hline
	&& \multicolumn{3}{c}{Threshold : 0.001 / 0.0001 / 0.00001} \\
	\cline{3-5}
	\multirow{7}{*}{AKAZE} & Raw & 26 / 130 / 714 & 13 / 52 / 147 & 11 / 50 / 139 \\
	& Linear & 26 / 130 / 715 & 13 / 53 / 148 & 11 / 50 / 141 \\
	& MBLLEN & 352 / 1055 / 1645 & 84 / 170 / \underline{\textbf{202}} & 81 / 160 / \underline{\textbf{188}} \\
	& GLAD & 256 / 1297 / 2191 & 59 / 146 / 171 & 57 / 137 / 160 \\
	& KinD & 177 / 1016 / 1871 & 54 / 157 / 194 & 52 / 150 / 182 \\
	& KinDpp & 457 / 1524 / 2127 & 80 / 143 / 156 & 78 / 136 / 146 \\
	& RetinexNet & 285 / 1661 / \underline{2614} & 37 / 87 / 97 & 35 / 81 / 90 \\
	\Xhline{2\arrayrulewidth}
	\hline
	&& \multicolumn{3}{c}{Quality Level : 0.01 / 0.001} \\
	\cline{3-5}
	\multirow{7}{*}{GFTT-BRIEF} & Raw & 113 / 522 & 49 / 115 & 42 / 106 \\
	& Linear & 126 / 598 & 55 / \underline{124} & 48 / \underline{116} \\
	& MBLLEN & 574 / 710 & 111 / \underline{124} & 102 / 113 \\
	& GLAD & 735 / 737 & 96 / 96 & 87 / 87 \\
	& KinD & 736 / \underline{746} & 123 / 123 & 112 / 113 \\
	& KinDpp & 730 / 730 & 109 / 109 & 101 / 101 \\
	& RetinexNet & 732 / 732 & 51 / 51 & 46 / 46 \\
	\Xhline{2\arrayrulewidth}
	\hline
	&& \multicolumn{3}{c}{Threshold : 30 / 10} \\
	\cline{3-5}
	\multirow{7}{*}{BRISK} & Raw & 72 / 296 & 23 / 68 & 20 / 65 \\
	& Linear & 72 / 296 & 23 / 68 & 21 / 65 \\
	& MBLLEN & 727 / 2624 & 74 / 155 & 70 / 145 \\
	& GLAD & 1049 / 5049 & 64 / 123 & 60 / 115 \\
	& KinD & 397 / 3112 & 63 / \underline{156} & 60 / \underline{147} \\
	& KinDpp & 1305 / 4940 & 86 / 142 & 83 / 131 \\
	& RetinexNet & 2287 / \underline{\textbf{7370}} & 43 / 67 & 40 / 58 \\
	\Xhline{3\arrayrulewidth}
\end{tabular}
\end{center}
\end{table*}

\begin{figure*}[ht]
    \centering
    \begin{tabular}{>{\centering\arraybackslash}m{0.05\linewidth} >{\centering\arraybackslash}m{0.445\linewidth} >{\centering\arraybackslash}m{0.445\linewidth}}
    SIFT &
    \includegraphics[width=1\linewidth]{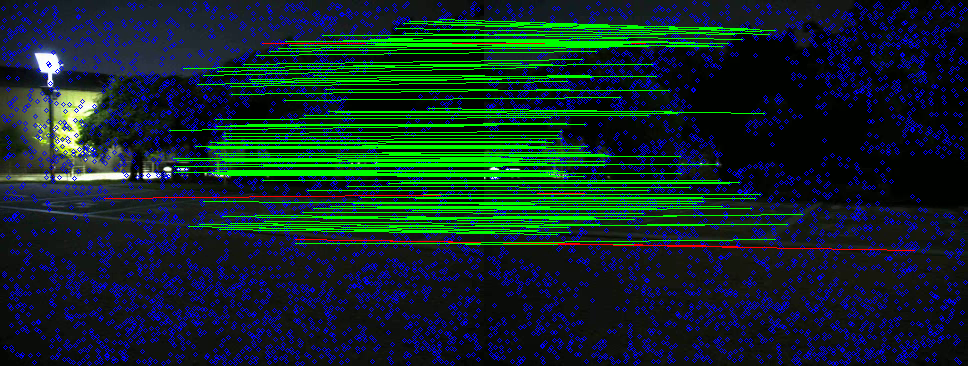} &
    \includegraphics[width=1\linewidth]{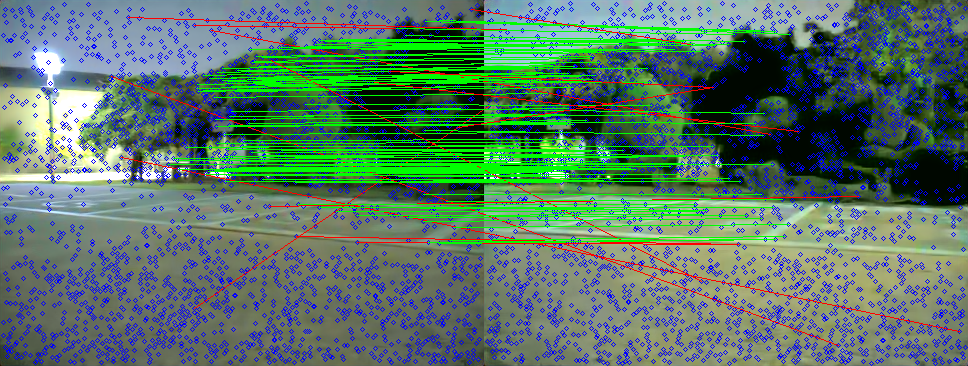} \\
    SURF &
    \includegraphics[width=1\linewidth]{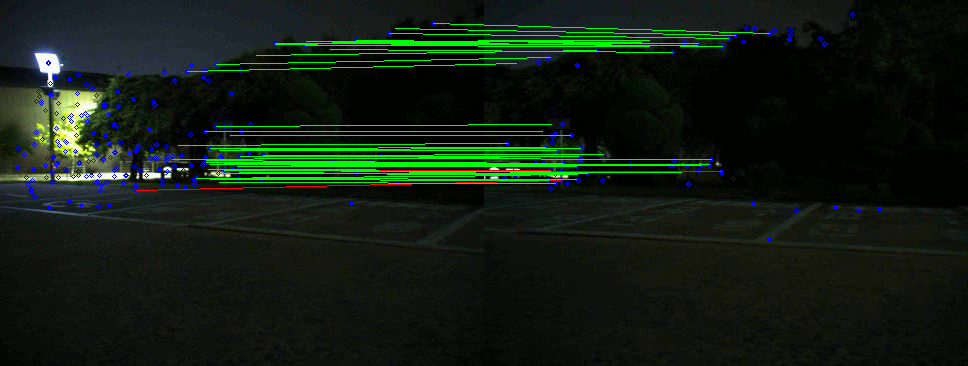} &
    \includegraphics[width=1\linewidth]{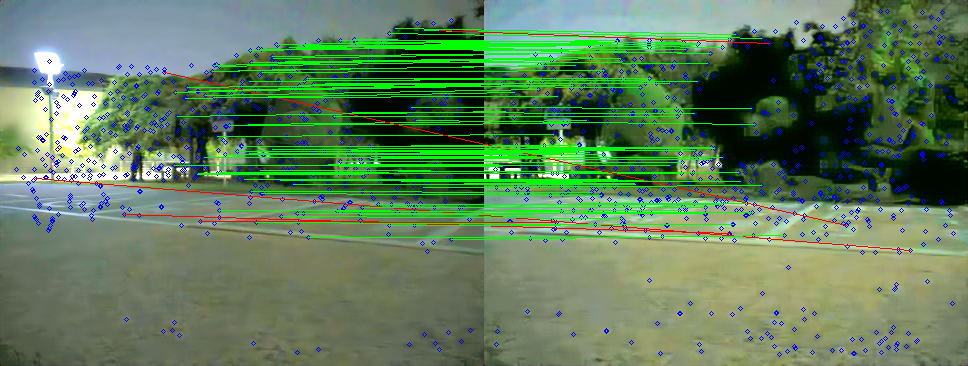} \\
    ORB &
    \includegraphics[width=1\linewidth]{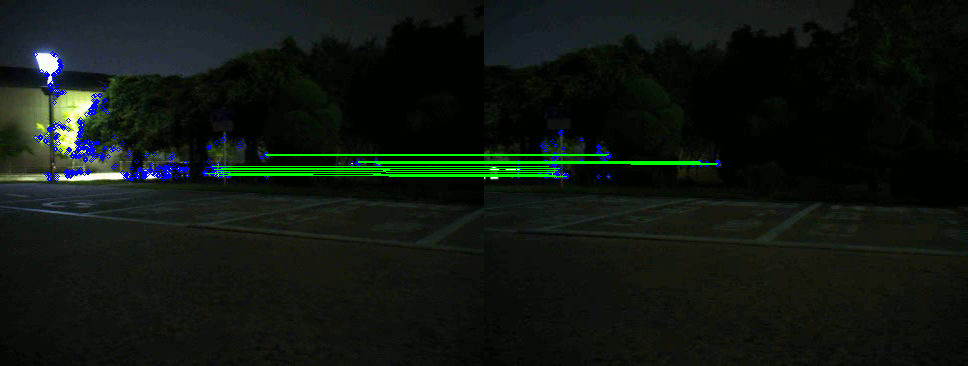} &
    \includegraphics[width=1\linewidth]{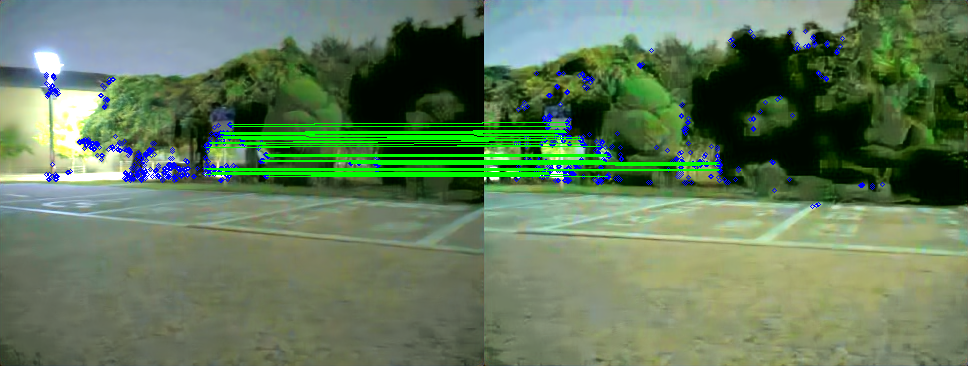} \\
    AKAZE &
    \includegraphics[width=1\linewidth]{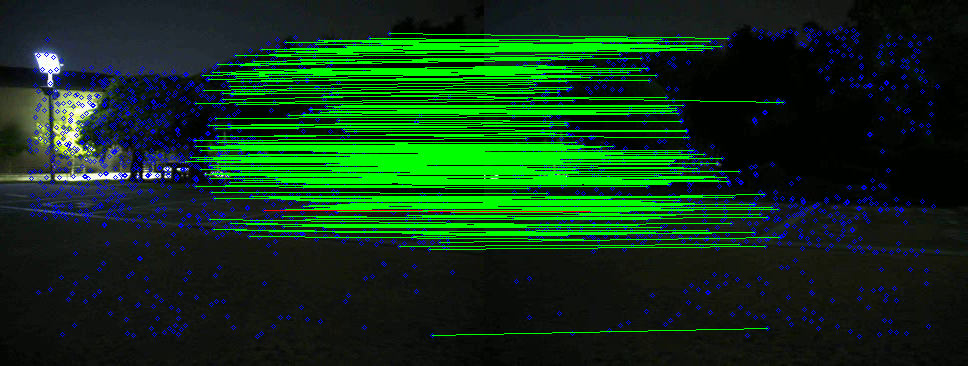} &
    \includegraphics[width=1\linewidth]{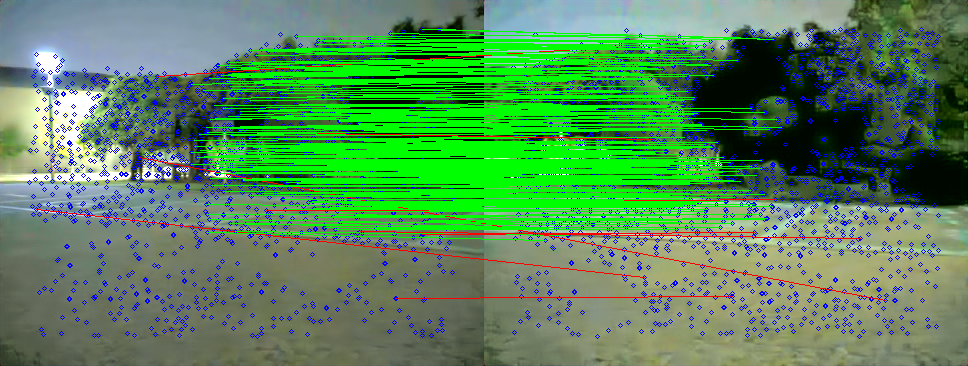} \\
    GFTT-BRIEF &
    \includegraphics[width=1\linewidth]{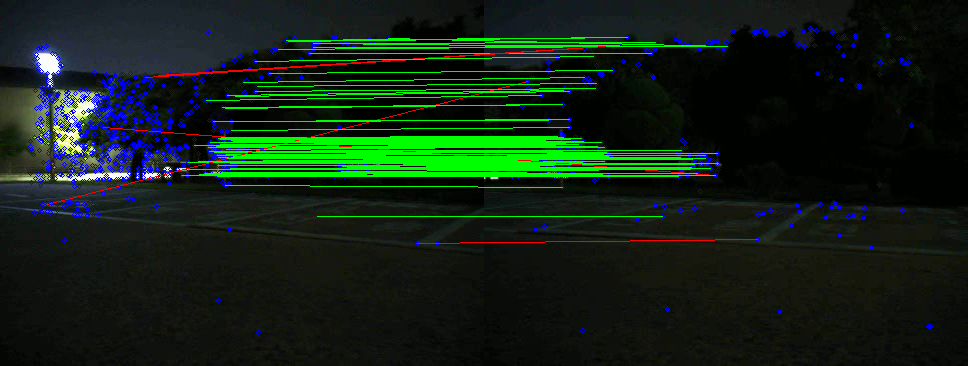} &
    \includegraphics[width=1\linewidth]{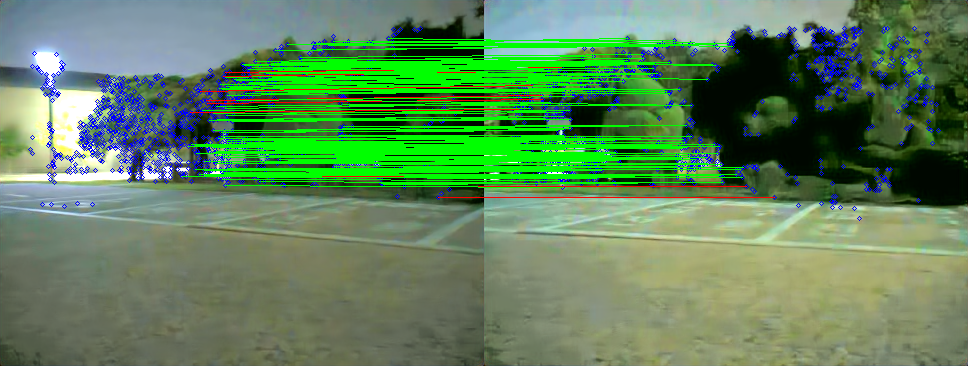} \\
    BRISK &
    \includegraphics[width=1\linewidth]{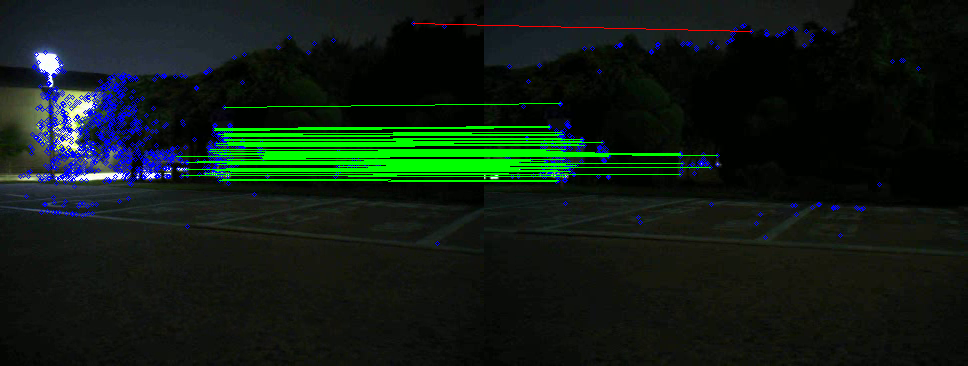} &
    \includegraphics[width=1\linewidth]{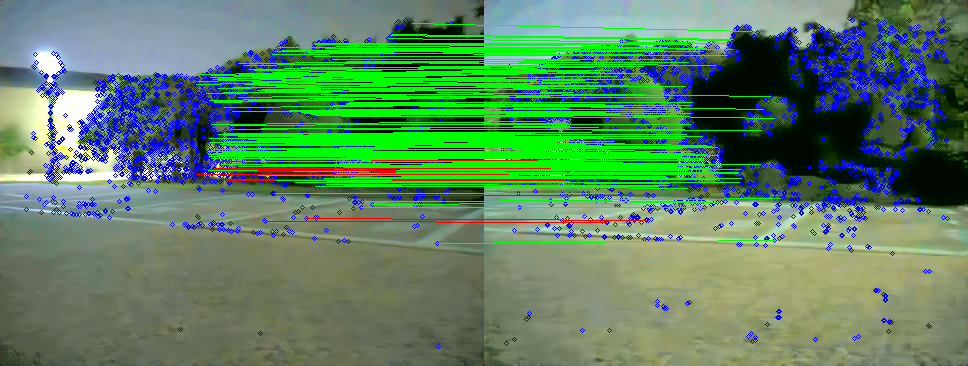} \\
    & Low Light Image Pair & MBLLEN Enhanced Image Pair
    \end{tabular}
    \captionsetup{justification=centering}
    \caption{Feature matching of image pairs using feature detector-descriptor at their best performing thresholds. Blue points are extracted features green lines represents inlier matches and red lines represents outlier matches.}
    \label{fig:matching}
\end{figure*}

\figref{fig:matching} shows the examples of feature matching results of different feature extractors on the raw images and MBLLEN processed images. The images shown are the matching results at the lowest feature detection threshold, hence the large number of features. 

\section{Acknowledgements}
This paper was supported by the Korea Civil-Military Combined Technology Development project (Task No. 19-CMGU-02).
\newpage
%%%%%%%%%%%%%%%%% BIBLIOGRAPHY IN THE LaTeX file !!!!! %%%%%%%%%%%%%%%%%%%%%%
%%---------------------------------------------------------------------------%%
%
\bibliographystyle{ieeetr}
\bibliography{egbib}
%
%%--------------------------------------------------------------------%%

\end{document}